\def\year{2022}\relax
\documentclass[letterpaper]{article} 
\usepackage{aaai22}  
\usepackage{times}  
\usepackage{helvet}  
\usepackage{courier}  
\usepackage[hyphens]{url}  
\usepackage{graphicx} 
\usepackage{caption} 
\DeclareCaptionStyle{ruled}{labelfont=normalfont,labelsep=colon,strut=off} 
\usepackage{algorithm}
\usepackage{algorithmic}
\usepackage{amsmath}
\usepackage{multirow}
\usepackage{seqsplit}

\usepackage{newfloat}
\usepackage{listings}
\usepackage{xcolor}
\usepackage{booktabs}
\usepackage{fancyhdr}
\floatstyle{ruled}
\newfloat{listing}{tb}{lst}{}
\floatname{listing}{Listing}

\title{Beyond the Black Box: Do More Complex Deep Learning Models Provide Superior XAI Explanations?}

\author {
    Mateusz Cedro\thanks{Corresponding author: mw.cedro@student.uw.edu.pl}\textsuperscript{\rm 1, \rm 2},
    Marcin Chlebus\textsuperscript{\rm 1}
}
\affiliations {
    \textsuperscript{\rm 1}University of Warsaw, Poland\\
    \textsuperscript{\rm 2}University of Antwerp, Belgium\\
}

\begin{document}

\maketitle

\begin{abstract}

The increasing complexity of Artificial Intelligence models poses challenges to interpretability, particularly in healthcare sector. This study investigates the impact of the deep learning models complexity and Explainable AI (XAI) efficacy, utilizing four ResNet architectures (ResNet-18, 34, 50, 101). Through methodical experimentation on 4,369 lung X-ray images of COVID-19-infected and healthy patients, the research evaluates models' classification performance and the relevance of corresponding XAI explanations with respect to the ground-truth disease masks. Results indicate that the increase in model complexity is associated with the decrease in classification accuracy and AUC-ROC scores (ResNet-18: 98.4\%, 0.997, ResNet-101: 95.9\%, 0.988). Notably, in eleven out of twelve statistical tests performed, no statistically significant differences occurred between XAI quantitative metrics - Relevance Rank Accuracy and proposed Positive Attribution Ratio - across trained models. These results suggest that increased model complexity does not consistently lead to higher performance or the relevance of explanations of models’ decision-making processes.
\end{abstract}
\raggedbottom
\section{Introduction}

Deep Neural Networks (DNNs) have garnered substantial success across diverse domains of Artificial Intelligence (AI) applications. Nonetheless, the opacity of their decision-making processes presents considerable challenges, particularly in sensitive sectors such as healthcare where transparency is essential \cite{morch_visualization_1995,baehrens_how_nodate,simonyan_deep_2014}. Efforts to reconcile the trade-off between model's accuracy and interpretability have led to the development of methods to trace predictions back to input features, enhancing model transparency \cite{lundberg_unified_2017,sundararajan_axiomatic_2017,arras_ground_2022,molnar2022}.

In the medical domain, the demand for interpretable models is heightened due to the life-or-death implications of decisions made \cite{rajkomar_machine_2019,holzinger_causability_2019}. Despite the intrinsic complexity of accurate machine learning models, medical experts require comprehensible insights into how specific features influence predictions \cite{che_interpretable_nodate,topol_high-performance_2019}.

Artificial neural networks, particularly DNNs, are designed to emulate the complexity of biological systems, resulting in architectures that are not inherently transparent, thus casting these models as "black-box" devices \cite{guidotti_survey_2019,ribeiro_why_2016,apley_visualizing_2020}. Consequently, enhancing model explainability is a key factor influencing the adoption of machine learning models in sensitive applications \cite{holzinger_what_2017}.

\subsection{Interpretable and Explainable AI}
The fields of Interpretable Machine Learning (IML) and Explainable Artificial Intelligence (XAI) have risen in response to the need for transparency in deep learning models, attracting significant attention in machine learning research \cite{ribeiro_why_2016, samek_explainable_2019, molnar_interpretable_2020}. Interpretability is defined by the ability of a machine learning system to make its processes and decisions understandable to humans. The quality of these explanations is crucial for model evaluation, validation, and debugging, and it's evaluated on the clarity of the model's decision-making process, not merely on prediction accuracy \cite{adebayo_sanity_2020}.

Interpretability is examined from two perspectives: global and local. Global interpretability provides an overarching understanding of the model's functioning and decision patterns, while local interpretability focuses on the specific rationale behind individual predictions. Both levels of interpretability are essential, serving varied purposes from enhancing scientific understanding to identifying biases and substantiating individual decisions \cite{doshi-velez_towards_2017, molnar2022}.

The challenge of explaining how neural networks arrive at their predictions is central to XAI, where the goal is to map and quantify the influence of each input feature on the final decision. This is particularly valuable in medical settings where healthcare professionals benefit from understanding the model's reasoning \cite{samek_explainable_2017,adebayo_sanity_2020}. Despite their capabilities, many machine learning models remain opaque, acting as "black boxes" without revealing the underlying logic guiding their decisions.

The attribution of a deep network's predictions to its input features has been identified as a fundamental challenge \cite{sundararajan_axiomatic_2017}. This attribution is represented as a vector, quantifying each input feature's contribution to the network's prediction, thereby clarifying the decision-making process, particularly beneficial for clinical experts in understanding the strengths and limitations of the model \cite{samek_explainable_2017, adebayo_sanity_2020}.

Techniques such as DeepLIFT \cite{shrikumar_learning_2019}, Layerwise Relevance Propagation \cite{bach_pixel-wise_2015}, LIME \cite{ribeiro_why_2016}, and Integrated Gradients \cite{sundararajan_axiomatic_2017} have been developed to unravel these decisions, breaking down the contributions of individual neurons to input features and thus advancing model interpretability. Moreover, in the GradientShap methodology, SHAP values (SHapley Additive exPlanations), have been adopted to attribute importance to each feature in a prediction, grounded in cooperative game theory, enhancing the interpretability of model features \cite{Shapley_1953,lundberg_unified_2017}.

Further refinements in interpretability methods, such as sensitivity and saliency maps, highlight influential image regions. Methods such as SmoothGrad and NoiseGrad have improved these techniques, reducing visual noise and integrating stochastic elements into models, improving both local and global interpretive clarity \cite{smilkov_smoothgrad_2017,bykov_noisegrad_2022}.

Evaluating feature significance within Explainable AI is essential for model optimization and establishing trust in model predictions \cite{hooker_benchmark_2019}. Challenges arise from the lack of universally accepted interpretability standards and the complexity involved in selecting and configuring appropriate interpretability methods \cite{kindermans_reliability_2017,arras_ground_2022}.

Evaluating feature significance often involves analyzing the effects of feature removal on model performance. This method, while effective, can alter the evaluation data distribution and thus potentially compromise the assessment's validity \cite{bach_pixel-wise_2015,sundararajan_axiomatic_2017}. With the growing necessity for transparent AI, objective and reproducible evaluation metrics are increasingly important \cite{ribeiro_why_2016}. One of the comprehensive frameworks for assessing the quality of the model explanations is Quantus \cite{hedstrom2023quantus}. This framework provides a comprehensive set of tools for accurate assessment of explanations and follows a transparent and impartial validation process for various XAI methodologies.

\subsection{AI and XAI in Medicine}
The integration of AI into healthcare is a strategic initiative aimed at personalizing patient treatment by harnessing the analytical prowess of AI to process and interpret large-scale clinical datasets \cite{lecun_deep_2015,holzinger_causability_2019}. Deep learning architectures, capable of sifting through extensive data such as hundreds of thousands of labelled X-ray images, are particularly instrumental in this shift from traditional rule-based diagnostics to a more nuanced, data-driven approach. This transition necessitates a framework within which the complex outputs of these models can be understood and trusted by medical professionals, a need met by the emerging field of Explainable AI \cite{cabitza_unintended_2017,rajkomar_machine_2019}.

XAI in medicine not only aims to unravel the decision-making processes of deep learning models but also strives to validate the reliability of AI-supported recommendations. The overarching goal is to establish a symbiotic relationship where AI systems are not merely tools for data extrapolation but partners in clinical decision-making, providing transparent and interpretable explanations that foster trust and facilitate informed medical judgments \cite{katuwal_machine_nodate, che_interpretable_nodate, hinton_deep_2018}.

AI carries transformative economic implications, necessitating a balance between peak performance and operational efficiency \cite{lecun_deep_2015, hinton_deep_2018, holzinger_causability_2019}. The deepening of neural networks, while advancing capabilities, approaches a threshold beyond which additional layers yield minimal performance gains, as identified by \cite{wu_wider_2016} and \cite{zhaounderstanding}. Contemporary advancements in complex architectural designs, such as Generative Pre-trained Transformers (GPT), have brought to the fore the significant financial and environmental costs inherent in the training processes. This development necessitates a judicious equilibrium between the advantages conferred by AI and the consumption of resources it entails \cite{brown_language_2020, menghani_efficient_2023}.

In healthcare, the role of AI is especially important as it offers the dual benefits of cost reduction and enhanced patient care. However, the adoption of AI must consider not just technological prowess but also the practicalities of application \cite{davenport_potential_2019,secinaro_role_2021}. This balance is essential in ensuring that AI's integration into healthcare remains both efficient and beneficial, providing clear, interpretable outcomes that align with the overarching goals of medical practice.

The COVID-19 pandemic has accelerated the application of deep learning computer vision models in medical diagnostics, as the disease can be identified on the X-ray images of infected patients' lungs. \cite{chowdhury_can_2020,degerli_covid-19_2021,rahman_exploring_2021}. Studies utilizing Residual Networks (ResNets) architectures \cite{he_deep_2015} on COVID-19 datasets have yielded promising results, underscoring the potential of deep learning in aiding pandemic response \cite{showkat_efficacy_2022}. Furthermore, the use of saliency maps in medical image segmentation has provided visual explanations that enhance the interpretability of model predictions, essential for medical diagnostics \cite{saporta_benchmarking_2022}.

\subsection{Influence of Model Scale on Performance and XAI Evaluations}
In the field of machine learning, there is a common hypothesis that an increase in model capacity should correlate with enhanced training efficacy \cite{eigen_understanding_2014}. Nonetheless, this correlation is not absolute, as studies have shown variable performance benefits with the scaling of model complexity, particularly in ResNet architectures \cite{dauphin_big_2013,wu_wider_2016}. While deeper neural networks such as ResNet-50, which is a 50-layer Convolutional Neural Network (CNN), have demonstrated improvements in specific tasks, these models do not universally outperform across all scenarios, with instances where less complex models like ResNet-18, an 18-layer CNN, match or exceed the accuracy of their larger counterparts \cite{khan_evaluating_2018, sarwinda_deep_2021}.

The concept of diminishing returns becomes evident as network complexity increases beyond a certain threshold, resulting in marginal performance enhancements that do not justify the additional complexity \cite{eigen_understanding_2014, wu_wider_2016}. In particular, ResNet-18 has been noted for its competitive performance against more elaborate models in certain classification tasks, prompting a reevaluation of the efficacy of scaling up network depth \cite{guo_classification_2019}. These observations underscore the imperative for a strategic approach in model selection that weighs computational efficiency against the specific performance requirements of the given task, thereby optimizing the balance between model architecture size and functional output.

To the best of our knowledge, no previous research has explored the relationship between the complexity of deep learning model architectures and the quality of XAI explanations. Our study is the first to address the problem in the literature by conducting experiments to investigate this relationship. In sectors where transparency is paramount, such as healthcare, understanding how architectural complexities affect both the model performance and the quality of XAI explanations is crucial. By conducting methodical experiments, this study aims to gain in-depth insight into the relationship between the complexity of deep learning models and the greatest possible interpretability, ultimately aiming to increase the accuracy and reliability of XAI explanations. Therefore, this study proposed two hypotheses.

\textbf{Hypothesis 1}: As the model's complexity increases, characterized by a greater number of trainable parameters,  it exhibits better classification performance.

\textbf{Hypothesis 2}: As the model's complexity increases, characterized by a greater number of trainable parameters, XAI assessment indicators are anticipated to yield inferior results, indicating an increased challenge in explaining the underlying decision-making process.

\section{Methodology}

To answer the underlying question, of whether more complex architectures provide better explainability in image classification tasks, in the conducted research the same workflow was employed for all of the trained ResNet models (ResNet-18, ResNet-34, ResNet-50, and ResNet-101). Initially, each ResNet model was trained from scratch, utilizing a consistent subset of randomly assigned images and model hyper-parameters to ensure equitable training conditions across all architectures.

After the training phase, a focused exploration into model explainability was undertaken by generating XAI explanations for each trained model, employing the Quantus library \cite{hedstrom2023quantus}. Three XAI methodologies were leveraged: Saliency Maps \cite{{morch_visualization_1995},{simonyan_deep_2014}}, GradientShap \cite{lundberg_unified_2017}, and Integrated Gradients \cite{sundararajan_axiomatic_2017}, each providing distinct perspectives into model decision-making processes.

The derived explanations were then subjected to a quantitative evaluation utilizing two pertinent metrics: Relevance Rank Accuracy \cite{arras_ground_2022} and proposed in this paper Positive Attribution Ratio, providing insightful revelations regarding the reliability and interpretability of the explanations propagated by each model. Having this approach, the following experiment provides a clear evaluation of the models' behaviour in the conducted image classification task.

\subsection{Data}

The dataset used in the following experiment was the COVID-QU-Ex dataset formulated by researchers from Qatar University and the University of Dhaka, which is a collection of the X-ray lung images obtained from various resources \cite{thair_COVID-QU-Ex}. The dataset contains three groups of X-rays: COVID-19 pneumonia, other diseases (non-covid), and healthy patients' lungs. For the X-rays from COVID-QU-Ex, corresponding ground-truth masks from the QaTa-COV19 dataset were used. QaTa-COV19 dataset was developed by Qatar University and Tampere University which provides binary segmentation masks of COVID-19 pneumonia \cite{QaTa-COV19_Database}.

For the following experiment, 4,369 X-ray lung images of different patients and corresponding ground truth masks were used. 2,913 images labelled as \textit{COVID-19} infected and 1,456 as \textit{Healthy}, non-infected patients. For training, validation and testing, X-ray images were randomly split on 70\%, 20\%, and 10\% dataset fractions respectively.

Before training, all the images were resized to the size of 224x224 pixels, turned into grayscale, transformed into tensors, and normalized. Transformations were done with the use of PyTorch's Torchvision library \cite{torchvision2016}.

\subsection{Models}
\begin{figure}[t]
\centering
\includegraphics[width=1\columnwidth]{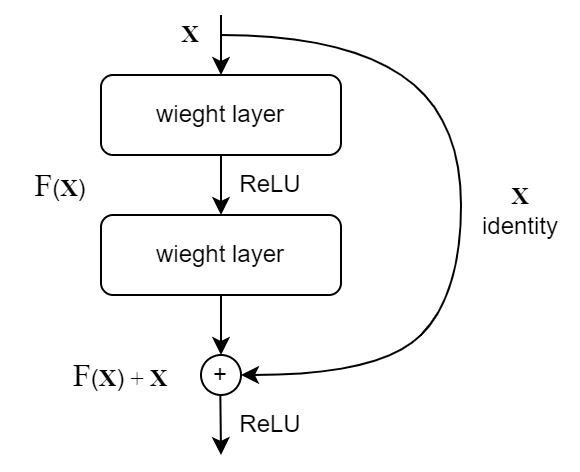}
\captionsetup{singlelinecheck=off, justification=raggedright}
\caption{Residual learning: a building block. 
\newline Source: Own preparation based on \cite{he_deep_2015}.}
\label{residual_connection}
\end{figure}

\begin{figure}
\centering
\includegraphics[width=0.75\columnwidth]{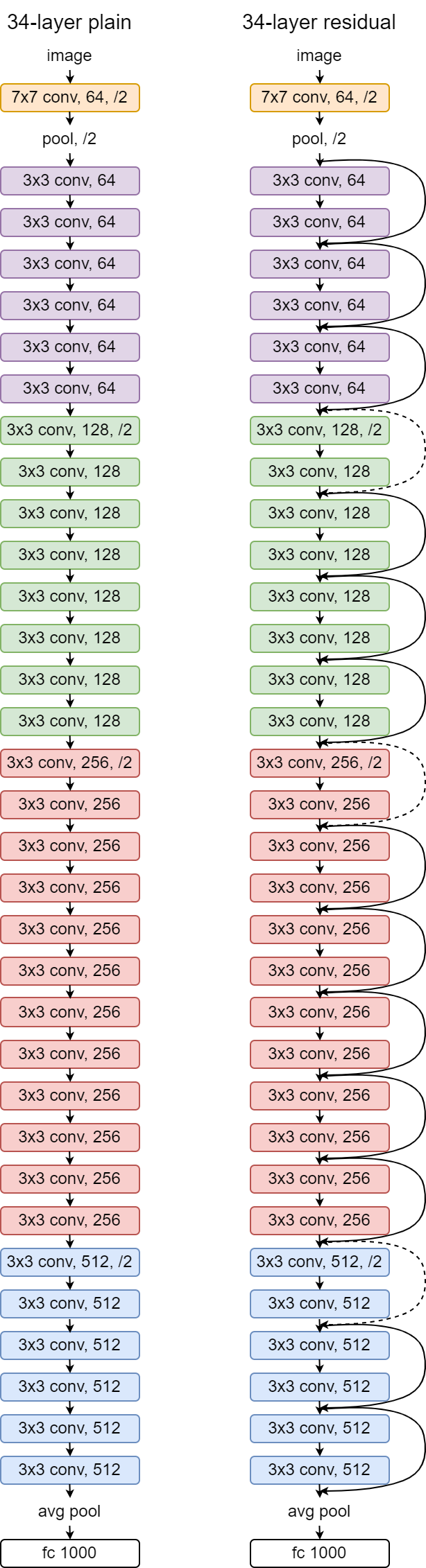}
\captionsetup{singlelinecheck=off}
\caption{Computer Vision network architectures. Left: 34-layer plain network. Right: 34-layer residual network. Dotted lines represent dimension-expanding connections.
\newline Source: Own preparation based on \cite{he_deep_2015}.}
\label{resnet34}
\end{figure}

In our experiment four ResNets architectures were explored, each distinguished by its depth: ResNet-18, ResNet-34, ResNet-50, and ResNet-101, where the suffix indicates the respective number of layers in the CNN models \cite{he_deep_2015}. Recognized for effectively addressing challenges in training deep networks for image classification tasks, these architectures were selected to probe the relationship between network depth and performance. Figure \ref{residual_connection} shows a building block containing the residual connection that provides an \textit{identity} input to every other layer, which became a state-of-the-art building block of deep learning architectures. Figure \ref{resnet34} presents a ResNet-34 architecture in comparison with plain 34-layer deep learning architecture \cite{he_deep_2015}.
We did not consider using pre-trained ResNet models because they are primarily trained on the ImageNet \cite{imagenet} dataset, which consists of natural images. These are significantly different from medical lung X-ray images, and hence would not improve the performance of medical image classification \cite{Towards_a_Better_Understanding_of_Transfer_Learning}. Additionally, we did not use the CheXNet \cite{chexnet} pre-trained model, which was trained on over 100,000 frontal-view X-ray images with 14 diseases, because it is a fixed-size model that bases on DenseNet-121 \cite{densely}, and there is no possibility to compare the XAI explanations to other CheXNet models, since there are no other pre-trained CheXNet models that base on a different number of DenseNet layers.

\begin{figure*}[t]
\includegraphics[scale=.5]{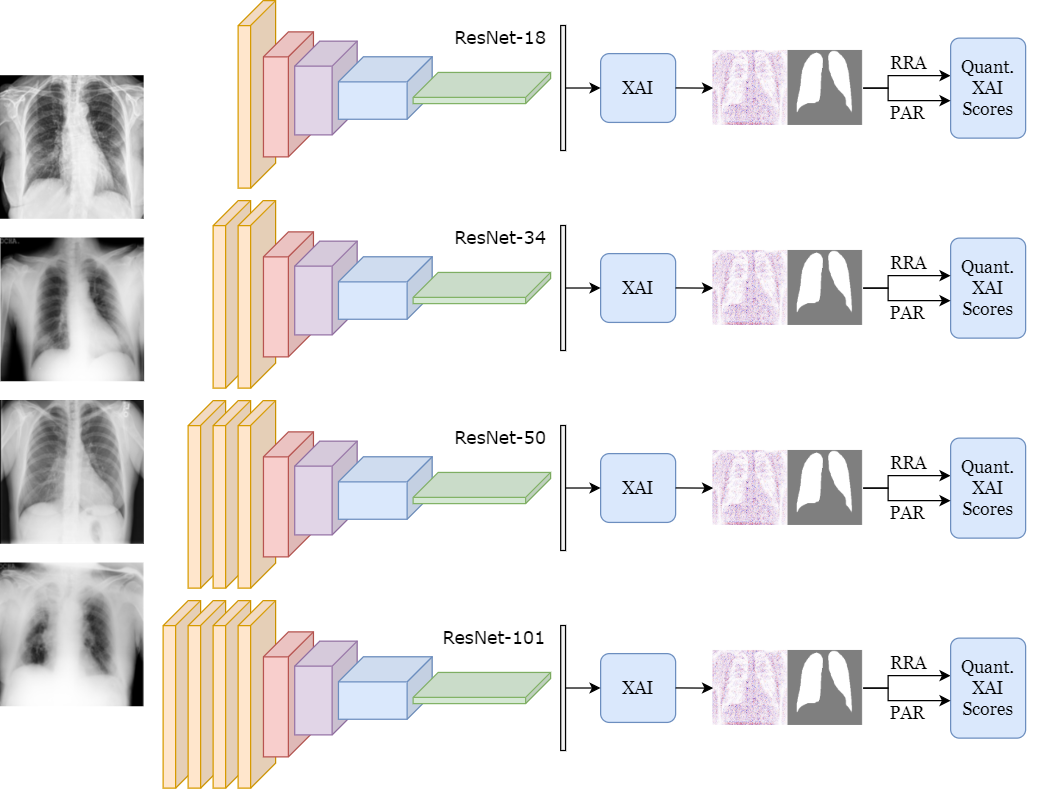}
\captionsetup{singlelinecheck=off}
\captionof{figure}{Comprehensive Workflow Schema of the Research Methodology. The abbreviations RRA and PAR stand for Relevance Rank Accuracy and Positive Attribution Ratio, respectively. Note: the model representation is illustrative and may not precisely reflect the original and specific number and type of layers of each model.
\newline Source: Own preparation.}
\label{workflow}
\end{figure*}

Baseline performance was established using ResNet-18 and ResNet-34, which were chosen for their balance of predictive power and computational efficiency. In contrast, ResNet-50 and ResNet-101 were scrutinized for potential accuracy improvements, despite their increased computational costs. A uniform training and testing process was applied to all models to ensure a fair comparison, and the trade-offs between model size, computational demand, and predictive accuracy were elucidated in the context of our research.

In Table \ref{tab:parameters}, the number of trainable parameters for all ResNet models is presented. Each consequent ResNet model has approximately double the number of the trainable parameters of the former model.

\begin{table}[t]
    \centering
    \begin{tabular}{lc}
        \toprule
        Model & \parbox{3.75 cm}{\centering Number of\\Trainable Parameters} \\
        \midrule
        ResNet-18 & 6,139,842 \\
        ResNet-34 & 12,329,218 \\
        ResNet-50 & 23,532,418 \\
        ResNet-101 & 42,550,658 \\
        \bottomrule
    \end{tabular}
    \caption{Number of Trainable Parameters in ResNet Models.
    \newline Source: Own calculations.}
    \label{tab:parameters}
\end{table}

\subsection{Model Training Setup}
In our research all ResNet architectures were trained from scratch for the image classification tasks. From the original dataset, the X-rays labelled as other diseases (non-covid) were excluded, leaving a dataset categorized into two label groups: \textit{COVID-19} and \textit{healthy}. Under the aforementioned approach, all models conducted binary classification tasks.

A concerted approach was employed to ensure the coherent training, validation, and testing of all models, with the images being randomly partitioned into respective subgroups comprising 70\%, 20\%, and 10\% of the data that contained in total 4,369 images subdivided into 2,913 and 1,456 images of \textit{COVID-19} and \textit{healthy} groups respectively.

The models were developed using the PyTorch library \cite{Paszke_PyTorch_An_Imperative_2019} and utilized a Cross-Entropy Loss criterion. This criterion computes the cross-entropy loss between predicted and target class labels, facilitating the models' learning from the logits.

The optimization of the model parameters was undertaken using Stochastic Gradient Descent \cite{zinkevich_parallelized_nodate} with a learning rate and momentum of 0.001 and 0.9, respectively. All models were subjected to the training for 50 epochs, with a batch size of 64, to gauge their efficacy in distinguishing between the defined label groups under consistent hyper-parameter settings \cite{bertrand_hyper_parameter_nodate}. Although each model was trained for 50 epochs, the final evaluation on the test set was conducted using the best-performing model checkpoint, which was selected based on the lowest validation loss encountered during the training process.

Ensuring experimental reproducibility and consistency across all training sessions, the random seeds for PyTorch and NumPy were fixed at a value of 42 \cite{chen_towards_2022}.

All model training sessions and subsequent Explainable AI analyses were conducted utilizing the Nvidia A100 GPU with 40 GB of RAM capacity.

\begin{figure*}[t]
\centering
\includegraphics[width=1\textwidth]{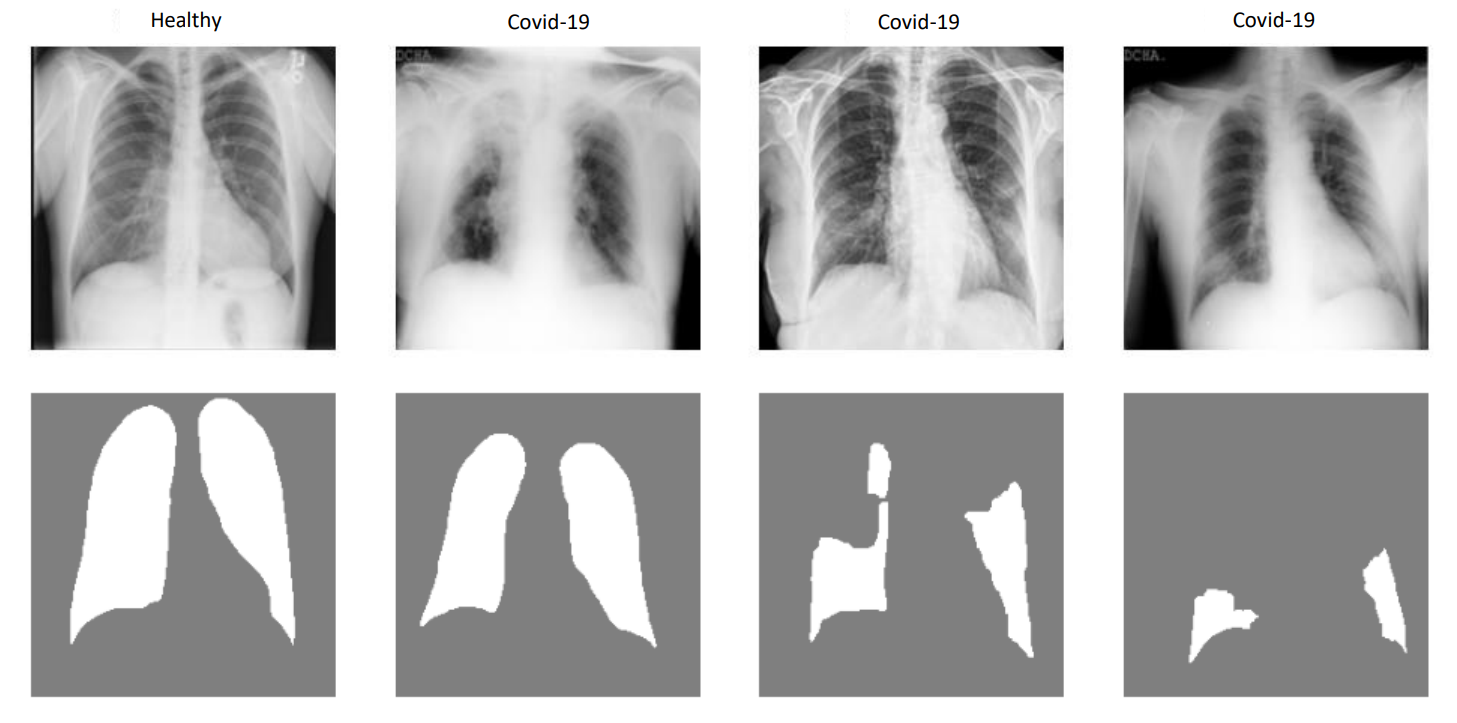}
\captionof{figure}{X-rays of \textit{healthy} and \textit{COVID-19} infected lungs, with corresponding ground-truth masks. Masks for healthy individuals encompass the entire lungs. In contrast, for COVID-19 patients, masks delineate areas identified by radiologists as diseased, potentially covering specific regions or the entirety of the lungs.
\newline Source: Own preparation.}
\label{pluca_maski}
\end{figure*}

\subsection{Gradient-based Techniques}

In the field of machine and deep learning, gradients are defined as the rate of change of the output with respect to the input and are acknowledged for their importance in the model's optimization. Formerly, the product of model coefficients with feature values has been examined by practitioners to interpret simpler, usually linear models. In deep neural networks, gradients are perceived as intrinsic coefficients, signifying the intricate connection between input and output \cite{{baehrens_how_nodate}, {simonyan_deep_2014}}. With advancements in research, gradient-based techniques have been introduced in the field of XAI, enabling a more profound interpretation of model behaviour and given prediction.

In this section, three gradient-based methods are outlined, specifically Saliency Maps \cite{{morch_visualization_1995},{baehrens_how_nodate},{simonyan_deep_2014}}, GradientShap \cite{lundberg_unified_2017}, and Integrated Gradients \cite{sundararajan_axiomatic_2017}. In the Saliency Maps method, the derivative of the class score with respect to the input image is calculated, identifying pixels that, when slightly altered, are found to have the most significant influence on the class score. Subsequently, the GradientShap method synthesizes Shapley values and gradients, further enhancing the understanding of model predictions. Lastly, the Integrated Gradients method is presented, wherein the path integration between input and output is detailed, providing a comprehensive attribution explanation.

A comprehensive examination of these gradient-based methodologies is undertaken in this chapter, highlighting their roles in augmenting the interpretability and transparency of deep neural architectures.

\begin{figure*}[t]
\centering
\includegraphics[width=1.1\textwidth]{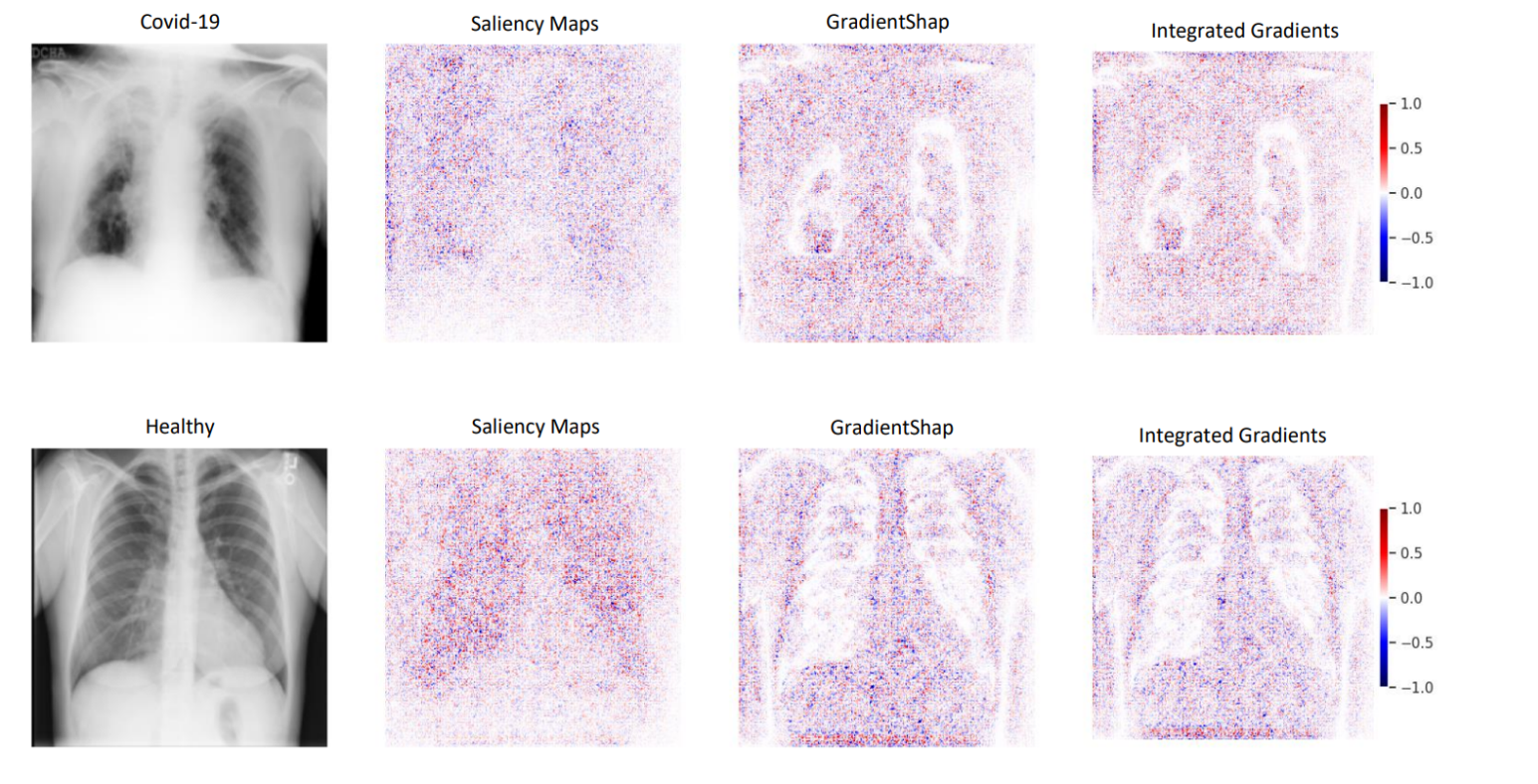}
\captionof{figure}{X-rays of \textit{healthy} and \textit{COVID-19}-infected lungs, accompanied by gradient-based attributions from Saliency Maps, GradientShap, and Integrated Gradients methodologies.
\newline Source: Own preparation with the use of Quantus library.}
\label{atrybucje}
\end{figure*}

X-rays of both healthy and COVID-19-infected lungs, along with their respective ground-truth masks and pixel attribution maps, are presented in Figure~\ref{pluca_maski} and Figure~\ref{atrybucje}.

\subsubsection{Saliency Maps}
One of the pioneering methodologies in the field of XAI is denoted as \textit{saliency maps} \cite{{morch_visualization_1995},{simonyan_deep_2014}}, which delineates the significance of specific components, such as pixels on the image, concerning the observed empirical relationships.

Given the inherent nonlinearity of the models with complex architecture, straightforward interpretations become elusive \cite{simonyan_deep_2014}. In this context, the saliency maps serve as an instrumental visualization mechanism, highlighting regions within the image that exhibit strong correlations to distinct tasks. By employing this technique, a transition from the high-dimensional input data space to a substantially reduced vector of projections is facilitated. This process inherently involves profound weight sharing, underlined by associations amongst weights interfacing the input and hidden layers of the feed-forward neural network architecture, as CNNs are. The saliency attributed to an input channel (for instance, the pixel \( i \) of an image vector) is quantified by the noticeable alteration in the cost function upon its exclusion.

In the research presented in \textit{Deep Inside Convolutional Networks: Visualising Image Classification Models and Saliency Maps} \cite{simonyan_deep_2014}, a gradient-based technique was introduced to compute an image-specific class saliency map, tailored to a distinct image and class combination. This method was harnessed using classification ConvNets and was designed to identify and highlight the spatial significance of a specific class within a given image. Essentially, for a given image \( I_0 \) and its associated class \( c \), the pixels of \( I_0 \) were ranked based on their impact on the class score \( S_c(I_0) \). Given the intricate non-linearity of deep ConvNets, the class score \( S_c(I) \) was approximated linearly in the vicinity of \( I_0 \) using the first-order Taylor expansion. Through this approach, pixels that could be altered minimally to most influence the class score were explained.

The procedure involved first determining the derivative \( w \) via back-propagation. Subsequently, the saliency map was extracted by reorganizing the components of vector \( w \). For grey-scale images, the dimensions of \( w \) were found to align with the pixel count of \( I_0 \), allowing for the map's computation as \( M_{ij} = |w_{h(i,j)}| \), where \( h(i, j) \) denoted the index of \( w \) that corresponded to the pixel situated in the i-th row and j-th column. Notably, this saliency map derivation utilized a classification ConvNet, trained exclusively on image labels, thereby eliminating the need for supplemental annotations, such as bounding boxes or segmentation masks.

To address the problem of accuracy-interpretability trade-off, \cite{lundberg_unified_2017} proposed an explanation framework named SHAP - SHapley Additive exPlanations. SHAP undertakes the model's feature interpretability on the concept from cooperative game theory \cite{Shapley_1953} by allocating an importance value to each feature for a specific prediction.

In research on model interpretability, it is commonly addressed that a simple model acts naturally as its own best explanation, eliminating the need for additional clarifications \cite{lundberg_unified_2017}. However, for complex models like deep neural network architectures, the original model is not interpretable by its nature. Thus, a more straightforward, interpretable model approximation or the \textit{explanation model} is needed. Consider denoting the original model as \( f \) and the \textit{explanation model} as \( g \). Explanation models typically employ simplified inputs, \( x_0 \), which correlate to the original inputs via a transformation function, \( x = h_x(x_0) \). The objective of local methods is to ensure that \( g(z_0) \) closely mirrors \( f(h_x(z_0)) \) whenever \( z_0 \) is akin to \( x_0 \).

\subsubsection{Shapley Values} In the field of cooperative game theory, the Shapley value is a fundamental mechanism designed to equitably allocate gains and costs among various participants within a coalition \cite{Shapley_1953}. This concept, originally formulated by Lloyd Shapley, becomes indispensable in scenarios where distinct actors contribute unequally yet collaborate towards a shared objective. The central premise of the Shapley value is to guarantee that each participant receives a payoff commensurate with their contribution, ensuring it is not less than what they would achieve independently. To clarify, within a strategic game involving multiple players aiming for a specific outcome, the Shapley value quantifies the average marginal contribution of each player, after considering all feasible combinations.

In a machine learning framework, the traditional players of the cooperative game are analogously represented by the features inherent to the machine learning model, with the model's output serving as a corollary to the game's payoff \cite{merrick_explanation_2019}. Shapley values offer a perspective on feature importance within linear models, particularly when multicollinearity is present. The application of this method necessitates the retraining of the model for all feature subsets \( S \subseteq F \), where \( F \) denotes the complete set of features. Each feature has assigned an importance value, representing its impact on the model prediction when included. To determine this impact, one model, \( f_{S \cup \{i\}} \), incorporates the particular feature, while the other, \( f_S \), excludes it. The predictions of these two models are subsequently contrasted based on the current input: \( f_{S \cup \{i\}}(x_{S \cup \{i\}}) - f_S(x_S) \), wherein \( x_S \) symbolizes the values of the input features contained within set \( S \). Given that the ramifications of omitting a feature are influenced by the model's other features, the aforementioned differences are evaluated across all feasible subsets \( S \subseteq F \setminus \{i\} \). Subsequent calculations yield the Shapley values, then formally, the contribution \( \phi \) of model feature \( i \) is defined as:

\begin{equation}
\phi_i = \sum_{S \subseteq F \setminus \{i\}} \frac{|S|!(|F| - |S| - 1)!}{|F|!} [ f_{S \cup \{i\}}(x_{S \cup \{i\}}) - f_S(x_S)]
\end{equation}

Conceptually, the Shapley value quantifies the average contribution of a specific feature \( i \), by evaluating the incremental payoff introduced by \( i \) across all possible coalitions that exclude feature \( i \).

\subsubsection{SHAP Values} SHAP values are proposed as a unified measure of feature importance, representing the Shapley values of a conditional expectation function of the original model \cite{lundberg_unified_2017}. These values are attributed to each feature, reflecting the change in the expected model prediction upon conditioning on that particular feature. The transition from the base value \( E[f (z)] \) — which would have been predicted in the absence of any known features — to the current output \( f (x) \) is explained by SHAP values. 

The unique additive feature importance measure that adheres to several properties is provided by SHAP values. These properties encompass:

\textit{Local accuracy} — ensuring that the explanation model \( g(x_0) \) corresponds with the original model \( f (x) \) when \( x = h_x(x_0) \);

\textit{Missingness} — where features with \( x_{i} = 0 \) are constrained to have no attributed impact;

\textit{Consistency} — which mandates that if a model's alteration causes a simplified input's contribution to either increase or remain unchanged irrespective of other inputs, the attribution of that input should not diminish.

Conditional expectations are utilized to define simplified inputs within these values. Inherent in the SHAP value definition is a simplified input mapping, denoted as \( h_x(z_0) = z_S \), where \( z_S \) contains missing values for features absent in set \( S \). Owing to most models' inability to process arbitrary patterns of missing input values, \( f (z_S) \) is approximated with \( E[f (z) | z_S] \). This definition of SHAP values is structured to closely resonate with the foundational Shapley values \cite{{Shapley_1953},{lundberg_unified_2017}}.

\subsubsection{GradientShap} GradientShap method estimates SHAP values by evaluating the gradient expectations, achieved by random sampling from a baseline distribution. By introducing white noise to input samples multiple times, it randomly selects a baseline and an intermediate point between the baseline and input, then calculates the gradient with respect to these random points. The resulting SHAP values mirror the expected values of these gradients multiplied by the difference between inputs and baselines.

The underlying assumption with GradientShap presumes that input features are independent and the explanation model is linear, indicating that the interpretations are modelled using the additive composition of feature effects. However, if the model exhibits non-linearity or the input features lack independence, the sequence in which features are incorporated into the expectation becomes significant. Under these circumstances, SHAP values are derived by averaging the Shapley values across all conceivable sequences. Given these conditions, the SHAP value can be approximated by the expected gradients computed for randomly generated samples, after Gaussian noise has been added to each input across various baselines.

\subsubsection{Integrated Gradients}
The problem addressed in \textit{Axiomatic Attribution for Deep Networks} publication \cite{sundararajan_axiomatic_2017} concerned the issue that many previous gradient base methods broke at least one of the two axioms that should always be satisfied in feature attribution methods, namely \textit{sensitivity} and \textit{implementation invariance} axioms. To address this problem, the \textit{Integrated Gradients} method was presented.

The Integrated Gradients approach has emerged as a notable solution in the field of deep neural network interpretation. Rooted from an axiomatic framework inspired by economics literature, Integrated Gradients seeks to fulfil both \textit{sensitivity} and \textit{implementation invariance} axioms. This ensures that the computed attributions are not just artefacts of the method but genuinely reflect the network's behaviour \cite{sundararajan_axiomatic_2017}.

An attribution technique adheres to the \textit{sensitivity} criterion when, for any input and baseline differing in just one feature with different predictions, the differing feature receives a non-zero attribution. consequently, if the deep network's function exhibits no mathematical dependence on a particular variable, that variable's attribution is always zero. In practical terms, the absence of sensitivity can lead to gradients predominantly concentrating on irrelevant features.

Within the context of neural networks, two architectures are deemed functionally equivalent when they produce consistent outputs across all given inputs, even if their internal implementations differ considerably. For attribution techniques, it is essential to adhere to the principle of \textit{implementation invariance}. This principle ensures that the attributions remain consistent for networks that are functionally equivalent, regardless of their distinct structures.

In the Integrated Gradients method, gradients are systematically integrated between a designated baseline, usually a black image, and the actual input image. This technique identifies the presence or absence of distinct features, thereby highlighting the significance of specific pixels or features within the contextual framework. It is commonly called a path-attribution technique \cite{molnar2022}. Critically, Integrated Gradients is deemed a \textit{complete} path-attribution approach. This implies that the cumulative relevance scores across all input features equate to the disparity between the prediction derived from the actual image and that of the reference image. In computer vision applications, pixel-wise attributions are presented, highlighting the areas of an image that resulted in the model's decision-making process.

\subsection{Evaluation Metrics}

In the context of machine learning interpretability, ensuring rigorous evaluation of explanatory heatmaps is crucial for computer vision models, especially when discerning model relevance. Therefore, to evaluate all the XAI approaches that have been used in this research study, \textit{Relevance Rank Accuracy} \cite{arras_ground_2022} and proposed in this paper \textit{Positive Attribution Ratio} metrics were used. Since both metrics are calculating the ratios,  their values fall within the [0-1] range, with a higher score signifying a more precise relevance heatmap.

\subsubsection{Relevance Rank Accuracy}

The Relevance Rank Accuracy is defined to gauge the degree to which the most pronounced relevance points are aligned with the ground truth. First, \( K \) is determined, representing the size of the ground truth mask. Then, the top \( K \) relevance values are extracted. Afterwards, the number of these values that correspond to locations within the ground truth is counted. This count is subsequently normalized by the dimension of the ground truth mask. Formally, this procedure can be expressed as:

\begin{equation}
P_{\text{top} K} = \{ p_1, p_2, \ldots, p_K \mid R_{p_1} > R_{p_2} > \ldots > R_{p_K} \}
\end{equation}

where \( P_{\text{top} K} \) represents the set of pixels, each associated with relevance values \( R_{p_1}, R_{p_2}, \ldots, R_{p_K} \), arranged in descending order up to the \( K \)-th pixel. Subsequently, the rank accuracy is determined as:

\begin{equation}
\text{Rank Accuracy} = \frac{|P_{\text{top} K} \cap \text{\( GT \)}|}{\text{\( |GT| \)}}
\end{equation}

where \( GT \) represents the set of pixel positions contained within the ground truth mask, and \( |GT| \) denotes the total count of pixels within this mask.

\subsubsection{Positive Attribution Ratio}

The Positive Attribution Ratio derives its foundation from the Relevance Mass Accuracy outlined by \cite{arras_ground_2022}. Nevertheless, a pivotal distinction exists, since it solely operates on pixels that possess positive attribution. We believe that for the future end-user, it is more important to be informed about the ratio of the number of pixels that have positive attribution localized inside the ground truth mask with respect to all positively attributed pixels on the investigated image.

The Relevance Mass Accuracy is calculated by dividing the aggregated sum of relevance values located within the ground truth mask by the total relevance values across the entire image. Essentially, this metric evaluates the proportion of the explanation method's "mass" attributed to the pixels within the ground truth. Positive Attribution Ratio operates in a similar manner, however, it focuses solely on pixels with positive attributions. As such, the Positive Attribution Ratio indicates the proportion of positive attributions within the ground truth mask \( R_{within} \) with respect to the positive attributions across the entire image \( R_{total} \). This might be formally represented as:

\begin{equation}
\text{Positive Attribution Ratio} = \frac{R_{\text{within}}}{R_{\text{total}}}
\end{equation}
where
\begin{equation}
R_{\text{within}} = \sum_{{k=1 \atop \substack{\text{s.t. } p_k \in \text{GT}}}}^{|\text{GT}|} R_{p_k},
\forall R_{p_k} > 0
\end{equation}
and
\begin{equation}
R_{\text{total}} = \sum_{{k=1}}^{N} R_{p_k}, \forall R_{p_k} > 0
\end{equation}

where \( R_{p_k} \) denotes the relevance value corresponding to pixel \( p_k \) which has positive relevance attribution, \( \text{GT} \) encompasses pixel locations present within the ground truth mask, \( |\text{GT}| \)  signifies the count of pixels within this mask, and \( N \) stands for the overall pixel count in the image. 

\section{Experiments and Results}

\subsection{Models' Performance}
Performance of the each ResNet models in terms of accuracy, AUC-ROC and Cross-Entropy Loss metrics on the separated test set is presented in Table \ref{tab:results}. The ResNet-18 architecture achieved the highest accuracy of 98.4\% and an AUC-ROC of 0.997, alongside maintaining the lowest cross-entropy loss of 0.066, misclassifying only 7 out of 437 X-ray images in the hold-out test set. Although all models demonstrated high accuracies and AUC-ROC values exceeding 95.9\% and 0.988 respectively, an inverse relationship was noted between model complexity and performance metrics, with ResNet-101 registering the lowest accuracy and AUC-ROC scores in the series. These findings are consistent with the results reported by \cite{guo_classification_2019}.

This evaluation underscores the need to consider the trade-off between model complexity and predictive performance in the selection of suitable deep learning architectures for image classification.

\begin{table}[h]
    \centering
    \begin{tabular}{lccc}
        \toprule
        Model & \parbox{1.5 cm}{\centering Accuracy\\(\%) ↑} & AUC-ROC ↑ & \parbox{2.1 cm}{\centering Cross-Entropy\\Loss ↓} \\
        \midrule
        ResNet-18 & \textbf{98.4} & \textbf{0.997} & \textbf{0.066} \\
        ResNet-34 & 97.3 & 0.996 & 0.097 \\
        ResNet-50 & 96.1 & 0.995 & 0.168 \\
        ResNet-101 & 95.9 & 0.988 & 0.153 \\
        \bottomrule
    \end{tabular}
    \caption{Performance Metrics of ResNet Models on Test Set. Bolded values indicate the best results for each metric.
    \newline Source: Own calculations.}
    \label{tab:results}
\end{table}

\subsection{Results}

The quantitative evaluations of all ResNet architectures, utilizing both the Relevance Rank Accuracy and Positive Attribution Ratio metrics, are presented in Table \ref{tab:relevance_rank_accuracy} and Table \ref{tab:positive_attribution_ratio} respectively. These evaluations incorporated the aforementioned XAI methodologies: Saliency Maps, GradientShap, and Integrated Gradients. The same interpretative methodologies were uniformly implemented across four ResNet models and evaluated independently on a test set consisting of 292 X-ray images labelled as \textit{COVID-19} class and 145 X-ray images of \textit{Healthy} class.

In the context of the \textit{COVID-19} class, clear fluctuations in performance indicators are evident. For the Relevance Rank Accuracy metric, ResNet-18 registered the highest mean score of 0.199 (\textit{SD=0.1}) when analyzed through the Saliency Maps approach. Conversely, the application of GradientShap and Integrated Gradients methodologies resulted in the highest scores of 0.118 (\textit{SD=0.09}) and 0.119 (\textit{SD=0.09}), respectively, which were attributed to the ResNet-101 architecture.

\begin{table}[t]
    \centering
    \begin{tabular}{clccc}
        \toprule
        Class & \parbox{1.1 cm}{\centering Model}
        & \parbox{1 cm}{\centering Saliency\\Maps\\Mean (SD)}
        & \parbox{1 cm}{\centering Gradient\\Shap\\Mean (SD)}
        & \parbox{1.5 cm}{\centering Integrated\\Gradients\\Mean (SD)} \\
        \midrule
        \multirow{8}{*}{\textit{COVID-19}} & ResNet-18 & \textbf{0.199} & 0.116 & 0.118 \\
        & & \textit{(0.10)} & \textit{(0.10)} & \textit{(0.10} \\
        & ResNet-34 & 0.198 & 0.117 & 0.115 \\
        & & \textit{(0.10)} & \textit{(0.10)} & \textit{(0.10)} \\
        & ResNet-50 & 0.191 & 0.115 & 0.116 \\
        & & \textit{(0.11)} & \textit{(0.10)} & \textit{(0.10)} \\
        & ResNet-101 & 0.175 & \textbf{0.118} & \textbf{0.119} \\
        & & \textit{(0.12)} & \textit{(0.09)} & \textit{(0.09)} \\
        \midrule
        \multirow{8}{*}{\textit{Healthy}} & ResNet-18 & 0.303 & 0.248 & \textbf{0.251} \\ 
        & & \textit{(0.05)} & \textit{(0.07)} & \textit{(0.07)} \\
        & ResNet-34 & \textbf{0.305} & \textbf{0.249} & 0.248 \\
        & & \textit{(0.05)} & \textit{(0.07)} & \textit{(0.07)} \\
        & ResNet-50 & 0.301 & 0.243 & 0.248 \\
        & & \textit{(0.05)} & \textit{(0.07)} & \textit{(0.07)} \\
        & ResNet-101 & 0.290 & 0.248 & 0.249 \\
        & & \textit{(0.06)} & \textit{(0.08)} & \textit{(0.07)} \\
        \bottomrule
    \end{tabular}
    \caption{Mean and Standard Deviation Scores for Relevance Rank Accuracy. Bolded values represent the best results within each methodology.
    \newline Source: Own calculations.}
    \label{tab:relevance_rank_accuracy}
\end{table}

\begin{table}[t]
    \centering
    \begin{tabular}{clccc}
        \toprule
        Class & \parbox{1.1 cm}{\centering Model}
        & \parbox{1 cm}{\centering Saliency\\Maps\\Mean (SD)}
        & \parbox{1 cm}{\centering Gradient\\Shap\\Mean (SD)}
        & \parbox{1.5 cm}{\centering Integrated\\Gradients\\Mean (SD)} \\
        \midrule
        \multirow{8}{*}{\textit{COVID-19}} & ResNet-18 & \textbf{0.186} & 0.117 & \textbf{0.120} \\
        & & \textit{(0.12)} & \textit{(0.10)} & \textit{(0.10)} \\
        & ResNet-34 & 0.185 & 0.118 & 0.118 \\
        & & \textit{(0.12)} & \textit{(0.10)} & \textit{(0.10)} \\
        & ResNet-50 & 0.182 & 0.116 & 0.118 \\
        & & \textit{(0.12)} & \textit{(0.10)} & \textit{(0.10)} \\
        & ResNet-101 & 0.169 & \textbf{0.120} & 0.119 \\
        & & \textit{(0.13)} & \textit{(0.10)} & \textit{(0.10)} \\
        \midrule
        \multirow{8}{*}{\textit{Healthy}} & ResNet-18 & 0.308 & 0.250 & \textbf{0.253} \\ 
        & & \textit{(0.07)} & \textit{(0.08)} & \textit{(0.08)} \\
        & ResNet-34 & \textbf{0.315} & 0.255 & 0.250 \\
        & & \textit{(0.07)} & \textit{(0.08)} & \textit{(0.08)} \\
        & ResNet-50 & 0.304 & 0.242 & 0.245 \\
        & & \textit{(0.07)} & \textit{(0.07)} & \textit{(0.07)} \\
        & ResNet-101 & 0.292 & \textbf{0.263} & 0.252 \\
        & & \textit{(0.08)} & \textit{(0.09)} & \textit{(0.08)} \\
        \bottomrule
    \end{tabular}
    \caption{Mean and Standard Deviation Scores for Positive Attribution Ratio. Bolded values represent the best results within each methodology. 
    \newline Source: Own calculations.}
    \label{tab:positive_attribution_ratio}
\end{table}

In the evaluation of the \textit{Healthy} class, the Relevance Rank Accuracy metric exposed varying performance paths. The ResNet-34 architecture, when interfaced with the Saliency Maps methodology, achieved a mean score of 0.305 (\textit{SD=0.05}). However, when subjected to the GradientShap and Integrated Gradients methodologies, mean scores of 0.249 (\textit{SD=0.07}) and 0.251 (\textit{SD=0.07}) were predominantly associated with ResNet-34 and ResNet-18 architectures, respectively.

\begin{figure*}[ht] 
\centering

\begin{minipage}{.5\textwidth}
\centering 
\includegraphics[width=0.877\textwidth]{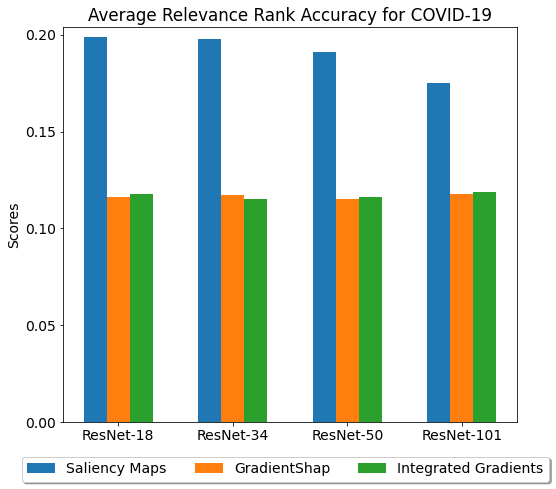}
\caption{Average results for relevance rank accuracy \newline within \textit{COVID-19} class.}
\label{rra_covid}
\end{minipage}%
\begin{minipage}{.5\textwidth}
\centering
\includegraphics[width=0.877\textwidth]{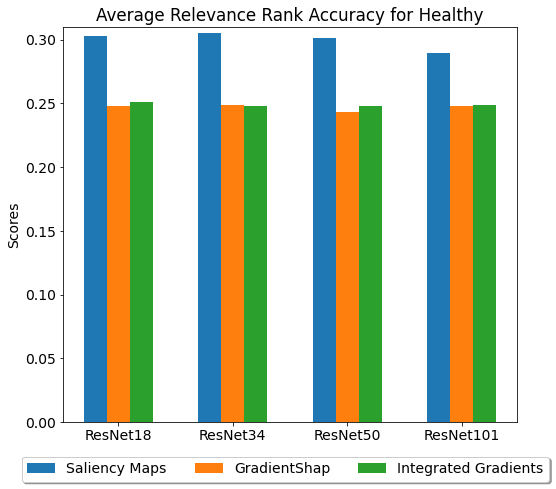}
\caption{Average results for relevance rank accuracy \newline within \textit{healthy} class.}
\label{rra_healthy}
\end{minipage}

\begin{minipage}{.5\textwidth}
\centering
\includegraphics[width=0.877\textwidth]{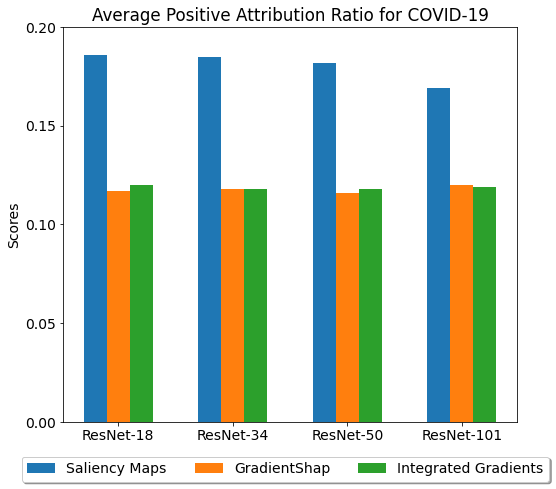}
\caption{Average results for positive attribution ratio \newline within \textit{COVID-19} class.}
\label{par_covid}
\end{minipage}%
\begin{minipage}{.5\textwidth}
\centering
\includegraphics[width=0.877\textwidth]{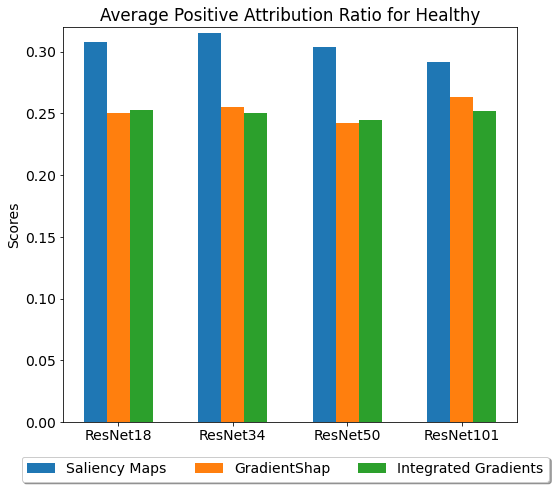}
\caption{Average results for positive attribution ratio \newline within \textit{healthy} class.}
\label{par_healthy}
\end{minipage}
\end{figure*}

Referring to the Positive Attribution Ratio scores within the \textit{COVID-19} group, ResNet-18 achieved the highest mean score of 0.186 (\textit{SD=0.12}) and 0.120 (\textit{SD=0.1}) within the Saliency Maps and  Integrated Gradients methodologies respectively. ResNet-101 achieved the highest mean score of 0.12 (\textit{SD=0.1}) in the GradientShap.

For the \textit{Healthy} class, under the Positive Attribution Ratio metric, ResNet-34 reached the highest mean score of 0.315 (\textit{SD=0.07}) using the Saliency Maps approach. In contrast, with the GradientShap methodology, ResNet-101 achieved a mean score of 0.263 (\textit{SD=0.09}), while ResNet-18 reached a mean score of 0.253 (\textit{SD=0.08}) with the Integrated Gradients approach.

\begin{table*}[t]
\centering
\setlength\tabcolsep{12pt} %
\begin{tabular}{lclcc}
\toprule
Metric & Class & Methodology & \parbox{2.2 cm}{\centering Kruskal-Wallis\\H-statistics} & \parbox{2.3 cm}{\centering Kruskal-Wallis\\\textit{p}-value} \\
\midrule
\multirow{6}{*}{\parbox{1.5cm}{Relevance\\Rank\\Accuracy}} &
\multirow{3}{*}{\textit{COVID-19}} & Saliency Maps & 9.79 & 0.02\rlap{\textsuperscript{*}} \\
& & GradientShap & 0.62 & 0.89 \\
& & Integrated Gradients & 1.19 & 0.76 \\
\cmidrule(lr){2-5}
& \multirow{3}{*}{\textit{Healthy}} & Saliency Maps & 5.83 & 0.12 \\
& & GradientShap & 0.58 & 0.90 \\
& & Integrated Gradients & 0.20 & 0.98 \\
\midrule
\multirow{6}{*}{\parbox{1.5cm}{Positive\\Attribution\\Ratio}} &
\multirow{3}{*}{\textit{COVID-19}} & Saliency Maps & 4.87 & 0.18 \\
& & GradientShap & 0.05 & 1 \\
& & Integrated Gradients & 0.30 & 0.96 \\
\cmidrule(lr){2-5}
& \multirow{3}{*}{\textit{Healthy}} & Saliency Maps & 7.37 & 0.06 \\
& & GradientShap & 3.90 & 0.27 \\
& & Integrated Gradients & 0.82 & 0.84 \\
\bottomrule
\end{tabular}
\captionsetup{singlelinecheck=off, justification=raggedright}
\caption{Statistical Test for Differences in Scores Across ResNet-18, ResNet-34, ResNet-50, and ResNet-101 Models.
\newline Note: * indicates statistical significance at the 0.05 level.
\newline Source: Own calculations.}
\label{tab:stat_scores_RRA}
\end{table*}

To assess the statistical significance of the differences between the means obtained from each of the ResNet architectures (18, 34, 50, and 101) for both metrics (Relevance Rank Accuracy and Positive Attribution Ratio), a procedure of statistical analyses was performed. These evaluations spanned across each of the three XAI methodologies (Saliency Maps, GradientShap, and Integrated Gradients) and were further separated based on two classes: \textit{COVID-19} and \textit{Healthy}. The analyses for the \textit{COVID-19} subgroup were conducted utilizing a set of 292 X-ray images, whereas the \textit{Healthy} class was assessed based on 145 X-rays. Both subgroups utilized images from the test set.

Upon analyzing the undertaken dataset, several observations emerge. Notably, ResNet-50 did not attain the top performance in either Relevance Rank Accuracy or Positive Attribution Ratio metric. Meanwhile, ResNet-18 secured the highest scores in five out of the twelve evaluated instances. ResNet-101 achieved the highest score in four out of the twelve instances, and lastly, ResNet-34 secured the top scores in three of the twelve evaluations. 

From the derived observations, it becomes clear that there is no direct correlation between the size or complexity of the ResNet model architecture and the resultant performance metrics like accuracy or AUC-ROC, aligning with the findings reported by \cite{khan_evaluating_2018, guo_classification_2019, sarwinda_deep_2021}. In terms of XAI quantitative metrics results, while ResNet-18 often displayed superior results, ResNet-50 did not necessarily follow suit, despite its increased complexity. Conversely, in certain scenarios, both ResNet-101 and ResNet-34 demonstrated superior performances, surpassing the results achieved by ResNet-50. Hence, it is imperative to understand that the choice of model architecture should not be solely based on its size or complexity. The results emphasize the importance of context-specific evaluations and suggest that in the domain of explainable AI for medical imaging, no one-size-fits-all approach is suitable.

Due to the unequal variances among groups, the Kruskal-Wallis test was used as a non-parametric alternative to the one-way ANOVA. This method evaluates the differences in medians across groups while accommodating the potential non-parametric distribution of the data, thus facilitating the detection of differences among the medians of the ResNet models. To clarify these differences, pairwise comparisons among the four ResNet models were performed using the Mann-Whitney U test. Considering the risk of type I errors due to multiple comparisons, the \( p \)-values obtained were adjusted using the Bonferroni correction method. With this statistical approach, a clear understanding of performance disparities across distinct model architectures with specific XAI techniques and image classes was achieved.

The statistical analysis reveals that the sole statistically significant divergence in medians across the ResNet architectures, at a threshold of \( p < 0.05 \), is discerned within the Relevance Rank Accuracy metric for the \textit{COVID-19} category, only for the Saliency Maps methodology (\( p = 0.02 \)). An extended analysis using the Mann-Whitney U test explained the statistically significant difference between the results for the ResNet-18 and ResNet-101 architectures, with \( p = 0.03 \). Furthermore, a marginal approach to significance within the same group and approach was observed between the ResNet-34 and ResNet-101 models, registering a \( p \)-value of 0.053.

In contrast, the remaining comparisons failed to evince any statistical discrepancies across the models, irrespective of the metric, category, or XAI technique in question. Notably, within the \textit{Healthy} category utilizing Saliency Maps, there was a boundary approach to statistical significance in the difference between medians across ResNet models, resulting in \( p \)-value of 0.06 for the Positive Attribution Ratio metrics.

\section{Discussion}
It is pertinent to note that the efficacy of interpretative methods in XAI hinges on their proper configuration \cite{montavon_explaining_2017, sundararajan_axiomatic_2017}. Incorrect settings can substantially diminish their effectiveness, as evidenced by past research \cite{kindermans_reliability_2017}. Therefore, constructing an empirical framework is crucial for validating the effectiveness and reliability of these methods \cite{hooker_benchmark_2019}. 

In healthcare and finance, users may mistakenly view predictive model outputs as causal, for instance, interpreting high saliency metrics as confirmation of specific health conditions. The capability of adversarial attacks to subtly alter inputs and shift focus from relevant to irrelevant features poses a significant challenge; such manipulations often go undetected as they do not change the diagnostic labels \cite{ghorbani_interpretation_2018}. The vulnerability of DNNs to these adversarial attacks is a documented concern, casting doubt on the trustworthiness of their predictive labels \cite{goodfellow_explaining_2015,kurakin_adversarial_2017,papernot_practical_2017}.

The research by \cite{ghorbani_interpretation_2018} delves into the impact of adversarial perturbations on the interpretations provided by neural networks. The interpretation of a neural network is considered vulnerable if there is a possibility to manipulate an image without a perceptual difference, maintaining the initial classification label, while significantly altering the network's interpretation of that image \cite{ghorbani_interpretation_2018}.

\section{Conclusions}
The influence of architectural complexity on the performance and explainability of ResNet models in medical image classification was investigated in this study. It was found that models with reduced complexity could deliver performance and interpretability comparable to or surpassing that of their more intricate counterparts. Specifically, architectures such as ResNet-18 were shown to provide effective accuracy and interpretability, challenging the prevailing belief that increased complexity ensures enhanced efficacy of the model. This provides grounds for \textbf{rejecting Hypothesis 1}.

Statistical analysis conducted on interpretability metrics on four ResNet models highlighted a lack of consistent correlation between architectural complexity and the quality of XAI explanations. The outcomes of this study necessitate a sensible approach to the selection of deep learning models, especially for applications that demand high precision and transparent explanations, such as those prevalent in healthcare. The results suggest that the additional resources required for more complex architectures, e.g. increased memory usage, higher financial costs, greater environmental impact, and longer training times, may not be justified, given that less complex architectures could achieve similar or superior levels of interpretability. This justifies the \textbf{rejection} of \textbf{Hypothesis 2}.

The study highlights the importance of properly configuring XAI methods to prevent misinterpretation of model predictions and urges for the development of an empirical framework to establish the reliability of these interpretive approaches. The conducted research reinforces the principle of a context-specific selection of neural network architectures underscoring the importance of both performance and interpretability, especially in applications within sensitive domains.

\section{Future Work}
In consideration of future explorations within the domain of Explainable Artificial Intelligence and image classification, it is crucial to address the growing interest in the Vision Transformer (ViT) architecture which surpasses the traditional CNN models in a variety of deep learning tasks \cite{vaswani_attention_2017, dosovitskiy_image_2020}. The inherent capacity of Transformers to facilitate complex, sequential data processing through self-attention mechanisms posits them as a prime candidate for augmenting the interpretability of deep learning models \cite{vaswani_attention_2017}.

Future investigations should strive to establish methodological approaches that quantify the effect of Vision Transformer complexity on explanation quality. This research should also extend to examining the capability of Transformers to preserve explainability when processing image datasets.

Additionally, it is crucial to extend the assessment of the explainability of Vision Transformers by leveraging the CheXpert \cite{chexpert} dataset, a comprehensive repository of 224,316 chest X-rays across 65,240 patients. The dataset encompasses 14 diverse radiological observations, each accompanied by annotations that mark uncertain diagnoses, providing a robust framework for appraising the interpretability of AI in the field of medical image analysis.

Such research endeavours are expected to contribute significantly to the development of AI systems that are both advanced in their operational capabilities and transparent in their reasoning processes. This balance is essential for building trust and enabling effective Human-AI interaction, propelling the field of XAI forward.

\section{Reproducibility}
The code utilized for replicating the experimental results is accessible at https://github.com/mateuszcedro/Beyond-the-Black-Box.

\bibliographystyle{unsrt}
\bibliography{Beyond-the-Black-Box}

\end{document}